\documentclass[conference, letterpaper]{IEEEtran}

\usepackage{amssymb}
\usepackage{algorithm}
\usepackage{algorithmic}
\usepackage{setspace}
\usepackage{flushend}
\usepackage{cite}
\usepackage{float}

\usepackage{subcaption}

\ifCLASSINFOpdf
   \usepackage[pdftex]{graphicx}
\else
\fi

\usepackage[cmex10]{amsmath}
\usepackage{color}
\usepackage{fancyhdr}

\renewcommand{\thispagestyle}[2]{} 

\fancypagestyle{plain}{
        \fancyhead{}
        \fancyhead[C]{first page center header}
        \fancyfoot{}
        \fancyfoot[C]{first page center footer}
}
\begin{document}

\title{LIUBoost : Locality Informed Underboosting for Imbalanced Data Classification}

\author{
	\IEEEauthorblockN{Sajid Ahmed, Farshid Rayhan, Asif Mahbub, Md. Rafsan Jani, \\Swakkhar Shatabda, Dewan Md. Farid and Chowdhury Mofizur Rahman}
	\IEEEauthorblockA{Department of Computer Science \& Engineering, United International University, Bangladesh\\  
		Email: dewanfarid@cse.uiu.ac.bd}}

\maketitle

\begin{abstract}

The problem of class imbalance along with class-overlapping has become a major issue in the domain of supervised learning. Most supervised learning algorithms assume equal cardinality of the classes under consideration while optimizing the cost function and this assumption does not hold true for imbalanced datasets which results in sub-optimal classification. Therefore, various approaches, such as undersampling, oversampling, cost-sensitive learning and ensemble based methods have been proposed for dealing with imbalanced datasets. However, undersampling suffers from information loss, oversampling suffers from increased runtime and potential overfitting while cost-sensitive methods suffer due to inadequately defined cost assignment schemes. In this paper, we propose a novel boosting based method called LIUBoost. LIUBoost uses under sampling for balancing the datasets in every boosting iteration like RUSBoost while incorporating a cost term for every instance based on their hardness into the weight update formula minimizing the information loss introduced by undersampling. LIUBoost has been extensively evaluated on 18 imbalanced datasets and the results indicate significant improvement over existing best performing method RUSBoost.                   

\end{abstract}


\IEEEpeerreviewmaketitle

\section{Introduction}
Class imbalance refers to the scenario where the number of instances from one class is significantly greater than that of another class. Traditional machine learning algorithms such as Support Vector Machines \cite{cortes1995support}, Artificial Neural Networks \cite{hopfield1988artificial}, Decision Tree \cite{safavian1991survey}, Random Forests \cite{breiman2001random} exhibit suboptimal performance when the dataset under consideration is imbalanced. This happens due to the fact that, these classifiers work under the assumption of equal cardinality between the underlying classes. However, many of the real world problems such as anomaly detection \cite{khreich2010iterative}, facial recognition \cite{liu2005total} where supervised learning is used are imbalanced. This is why researchers came up with different methods that would make the existing classifiers competent in dealing with classification problems that exhibit class imbalance.
 
 Most of these proposed methods can be categorized into sampling techniques, cost-sensitive methods and ensemble based methods. The sampling techniques either increase the number of minority class instances(oversampling) or decrease the number of majority class instances(undersampling) so that imbalance ratio decreases and the training data fed to some classifier becomes somewhat balanced \cite{studyofsamplingmethods}. The cost sensitive methods assign higher misclassification cost to the minority class instances which is further incorporated into the cost function to be minimized by the underlying classifier. The integration of these cost terms minimizes the classifiers' bias towards the majority class and puts greater emphasis on the appropriate learning of the minority concept \cite{costsensitiveboostingforimbalanceddata}. Ensemble methods such as Bagging \cite{breiman1996bagging} and Boosting \cite{adaboostfirstpaper} employ multiple instances of the base classifier and combine their learning to predict the dependent variable. Sampling techniques or cost terms are incorporated into ensemble methods for dealing with the problem of class imbalance and these methods have shown tremendous success \cite{rusboost,chawla2003smoteboost}. As a matter of fact, these ensemble methods turned out to be the most successful ones for dealing with imbalanced datasets \cite{galar2012review}.               
 
 In order to reduce the effect of class imbalance, the aforementioned methods usually attempt to increase the identification rate for the minority class and decrease the number of false negatives. In the process of doing so, they often end up decreasing the recognition rate of the majority class which results in a large number of false positives. This can be equally undesirable in many real world problems such as fraud detection where identifying a genuine customer as fraud could result in loss of loyal clients. This increased false positive rate could be due to under-representation of the majority class(undersampling), over-emphasized representation of the minority class(oversampling) or over-optimistic cost assignment(cost-sensitive methods). The most successful ensemble based methods also suffer from such problems because they use undersampling or oversampling for the purpose of data balancing while the cost-sensitive methods suffer from over-optimistic cost assignment because the proposed assignment schemes only take into account the global between-class imbalance and do not consider the significant characteristics of the individual instances \cite{sun2015novel}.   
 
 In this study, we propose a novel boosting based approach called \textbf{L}ocality \textbf{I}nformed \textbf{U}nder\textbf{boost}ing (LIUBoost) for dealing with class imbalance. The aforementioned methods have incorporated either sampling or cost-terms into boosting for mitigating the effect of class imbalance and have fallen victim to either information loss or unstable cost assignment. However, LIUBoost uses undersampling for balancing the datasets while retaining significant information about the local characteristics of each of the instances and incorporates that information into the weight update equation of AdaBoost in the form of cost terms. These cost terms minimize the effect of information loss introduced by undersampling. We have used K-Nearest Neighbor (KNN) algorithm \cite{mani2003knn} with small \textit{K} value for locality analysis and weight calculation. These weights are not meant to mitigate the effect of class imbalance in any way. However, these weights are able to differentiate among safe, borderline and outlier instances of both majority and minority classes and provide the underlying base learners with a better representation of both majority and minority concepts. Additionally, LIUBoost takes into account problems such as class overlapping \cite{garcia2007empirical}, the curse of bad minority hubs \cite{tomavsev2013class} that occur together with the problem of class imbalance. The aim of this study is to show the effectiveness of our proposed LIUBoost both theoretically and  experimentally. To do so, we have compared the performance of LIUBoost with that of RUSBoost on 18 standard benchmark imbalanced datasets and the results shows LIUBoost significantly improves over RUSBoost.


The remainder of the paper has been arranged as follows. Section~\ref{related_work} presents related work and motivation behind our proposal, Section~\ref{proposed_methods} presents our proposed method and Section~\ref{experiments} provides the experimental results. Finally, we conclude in Section \ref{conclusion}. 

\section{Related work}
\label{related_work}


Seiffert et al.\cite{rusboost} proposed RUSBoost for the task of imbalanced classification. RUSBoost integrates random under-sampling at each iteration of AdaBoost \cite{adaboostfirstpaper}. In different studies, RUSBoost has stood out as one of the best  performing boosting based methods alongside SMOTEBoost for imbalanced data classification \cite{galar2012review,khoshgoftaar2011comparing}. A major key to the success of RUSBoost is its random under-sampling technique which, in spite of being a simple non-heuristic approach, has been shown to outperform other intelligent ones \cite{randomundersamplingbetter}. Due to the use of this time-efficient yet effective  sampling strategy, RUSBoost is more suitable for practical use compared to SMOTEBoost \cite{chawla2003smoteboost} and other boosting based imbalanced classification methods which employ intelligent under-sampling or over-sampling, thus making the whole classification process much more time-consuming. However, RUSBoost may fall victim to information loss when faced with highly imbalanced datasets. This happens due to its component random under-sampling \cite{randomundersamplinginformationloss} which discards a large number of majority class instances at each iteration, thus the majority class is often underrepresented in the modified training data fed to the base learners. Our proposed method incorporates significant information about each of the instances of the unmodified training set into the iterations of RUSBoost in the form of cost in order to mitigate the aforementioned information loss.

 
 Fan et al. proposed AdaCost \cite{adacost} which introduced misclassification costs for instances into the weight update equation of AdaBoost. They theoretically proved that  introducing costs in this way does not break the conjecture of AdaBoost. However, they did not develop any generic weight assignment scheme that could be followed for different datasets. Their weight assignments were rather domain specific. Karakoulas et al. \cite{karakoulas1999optimizing} proposed a weight assignment scheme for dealing with the problem of class imbalance where false negatives were assigned higher weights compared to false positives. Sun et al. proposed three cost-sensitive boosting methods for the classification of imbalanced datasets AdaC1, AdaC2 and AdaC3 \cite{costsensitiveboostingforimbalanceddata}. These methods assign greater misclassification cost to the instances of the minority class. If an instance of the minority class is misclassified, its weight is increased more forcefully compared to a misclassified majority class instance. Furthermore, if a minority instance is correctly classified, its weight is decreased less forcefully compared to a correctly classified majority instance. As a result, appropriate learning of the minority instances is given greater emphasis in the training process of AdaBoost in order to mitigate the effect of class imbalance. All these methods assign an equal cost to all instances of the same class considering the between-class imbalance ratio. None of them take into account local characteristics of the data points.              
 

Most of the methods proposed for classification of imbalanced datasets only take into account the difference between number of instances from the majority and the minority class and try to mitigate the effects of this imbalance. However, this difference is only one of the several factors that make the task of classification extremely difficult. But these additional yet extremely significant factors are often overlooked while designing algorithms for imbalanced classification \cite{napierala2016types}. One of these factors is the overlapping of majority and minority classes. Prati et al. \cite{prati2004class} studied the effect of class overlapping combined with class imbalance by varying their respective degree and deduced that overlapping is even more detrimental to the classifier performance. Garcia et al. \cite{garcia2007empirical} examined the performance of six classifiers on datasets where class imbalance and overlapping was high and noticed that KNN \cite{mani2003knn} with a small value of \textbf{K}(local neighborhood analysis) was the best performer under such circumstances. These observations point towards the feasibility of dealing with the problem of class overlapping in imbalanced datasets through incorporating information about the local neighborhood of the instances into the training process. Another factor responsible for degrading the performance of classifiers in imbalanced datasets is the effect of bad minority hubs. These are instances of the minority class that are closely grouped together in the feature space. If such a group is close to a majority instance, that majority instance will have a high probability of being misclassified \cite{tomavsev2013class}. Such effects are not taken into account in the cost assignment scheme proposed by aforementioned cost-sensitive methods for imbalanced classification. However, our proposed method attempts to mitigate the effects of class-overlapping and bad minority hubs by taking into account the local neighborhood of each of the instances while assigning weights to them.                   


In some recent proposals, authors have incorporated locality information of the instances into their methods in different ways for dealing with imbalanced datasets. He et al. proposed ADASYN \cite{adasynpaper} over-sampling technique which takes into account number of majority class instances around the existing minority instances and creates more synthetic samples for the ones with more majority neighbors so that the harder minority instances get more emphasis in the learning process. Blaszczynski et al. proposed Local-and-Over-All Balanced Bagging \cite{blaszczynski2013extending} which integrates locality information of the majority instances into UnderBagging. In this approach, the majority instances with less number of minority instances in their local neighborhood are more likely to be selected in the bagging iterations. Bunkhumpornpat et al. proposed Safe-Level-SMOTE \cite{safelevelsmote} which only uses the safe minority instances for generating synthetic minority samples. Han et al. proposed Borderline-SMOTE \cite{borderlinesmote} which only uses the borderline minority instances for synthetic minority generation. Furthermore, Napierala et al. used locality information of the minority instances to divide them into aforementioned categories such as safe,borderline,rare and outlier \cite{napierala2016types}. All these aforementioned methods suggest that locality information of minority and majority instances is significant and can be used in the learning process of classifiers designed for imbalanced classification.    
\section{Proposed Method}
\label{proposed_methods}
\begin{algorithm}[H]
	\caption{Weight\_Assignment(dataset $X$,$k$)}
	\label{weight_assignment}
	\begin{algorithmic}[1]
		
		\FOR{each instance $x_i \in $ in the training set, $X$}	
		\STATE find \textbf{k} nearest nearest neighbors for the instance 
		\STATE $N_s \gets $ number of neighbors with same class
		\STATE $N_o \gets $ number of neighbors with opposite class
		\IF {$N_s == 0$}
		\STATE  $Weight^+(i) \gets \delta$  
		\STATE  $Weight^-(i) \gets \frac{1}{N_o}$ 
		\ELSIF { $N_o == 0$
		}				  
		\STATE  $Weight^+(i) \gets \frac{1}{N_s}$ 
		\STATE  $Weight^-(i)\gets \delta$ 
		\ELSE
		\STATE  $Weight^+(i)\gets \frac{1}{N_s}$ 
		\STATE  $Weight^-(i) \gets \frac{1}{N_o}$ 
		\ENDIF 
		\ENDFOR
		
		Return the $Weight^+$ and $Weight^-$
	\end{algorithmic}
\end{algorithm}
    
\begin{algorithm}[H]
	\begin{algorithmic}[1]
		\STATE $m$ = number of instances
		\STATE $T$ = number of boosting iterations
		\STATE $(Weight^+,Weight^-)$ = Weight\_Assignment($dataset$,$k$)
		\FOR{ $i\gets 1$ to $m$ }
			\STATE $D_i\gets \frac{1}{m}$
		\ENDFOR
		\FOR{ $t\gets 1$ to $T$ }
			\STATE $undersampled\_dataset$ = Undersampling($dataset$)
			\STATE $h_t \gets $ Decision\_Tree($undersampled\_dataset$) 
			\STATE  $mis\_sum \gets \sum_{
				\substack{h_t(x_i)\neq y_i}}D_i^t \cdot Weight^+(i)$
			\STATE  $cor\_sum \gets \sum_{
				\substack{h_t(x_i) = y_i}}D_i^t \cdot Weight^-(i)$ 
			\STATE update parameter $\alpha_t	\gets \frac{1}{2}\text{log}\frac{1+cor\_sum-mis\_sum}{1-cor\_sum+mis\_sum}$
			\IF{$\alpha_t \leq$ 0}
				\STATE  $t\gets t-1$
				\STATE return to statement 8 
			\ENDIF 
			\FOR{$i\gets 1$ to $m$}
				\IF{$y_i$ $\neq$ $h_t(x_i)$}
					\STATE $D_{t+1}(i) \gets D_t e^ {-\alpha_t \cdot y_i \cdot h_t(x_i) \cdot Weight^+(i)}$
				\ELSE 
					\STATE $D_{t+1}(i) \gets D_t e^ {-\alpha_t \cdot y_i \cdot h_t(x_i) \cdot Weight^-(i)}$
				\ENDIF
			\ENDFOR
			\STATE normalize $D$
		\ENDFOR
		\STATE $g \gets \sum_{
		\substack{t=1 
		}}^{T}\alpha_th_t$
		\STATE Return $h=sign(g)$
	\end{algorithmic}
	\caption{LIUBoost($dataset = (X,Y)$)}
	\label{LIUBoost}
\end{algorithm}
The pseudo code of our proposed method LIUBoost is given in Algorithm~\ref{LIUBoost}. LIUBoost calls Weight\_Assignment method given in Algorithm~\ref{weight_assignment} before boosting iterations begin. This method returns two sets of weights $Weight^-$ and $Weight^+$ used respectively to decrease and increase the weights associated an instance. $Weight^+$ are added inside the exponent term of the weight update equation for the misclassified instances at the iteration under consideration while $Weight^-$ are added for the correctly classified instances. As a result, weight of the instances with greater $Weight^+$ grow rapidly if they are misclassified while weight of the instances with greater $Weight^-$ drop rapidly if they are correctly classified. Thus LIUBoost puts greater emphasis on learning the important concepts rapidly. Additionally, LIUBoost performs undersampling at each boosting iteration for balancing the training set.

The alpha terms determine how significant the predictions of each of the individual  base learners are in the final voted classification. These terms also play an important role in the weight update formula which ultimately minimizes the combined error. Since LIUBoost has modified the original weight update equation of AdaBoost by adding cost-terms, the alpha term needs to be updated accordingly in order to preserve coherence of the learning process. The alpha term has been updated according to the recommendations from \cite{costsensitiveboostingforimbalanceddata}.

One thing to notice here is that LIUBoost combines sampling method and cost-sensitive learning in a novel way. The proposed weight assignment method assigns greater $Increase\_Weight$ to borderline and rare instances while assigning less $Increase\_Weight$ to safe instances due to the way it analysis local neighborhood. Napierala et al. \cite{napierala2016types} proposed a similar method for grouping only the minority instances into four categories such as safe, borderline, rare and outlier. However, LIUBoost also distinguishes the majority instances through weight assignment. When the majority and minority classes are highly overlapped, which is often the case with highly imbalanced datasets \cite{prati2004class}, undersampling may discard a large number of borderline and rare majority instances which will increase their misclassification probability. LIUBoost overcomes this problem by keeping track of such majority instances through assigned weights and puts greater emphasis on their learning. This is its unique feature for minimizing information loss.


\section{Experimental Results}
\label{experiments}
This section presents the details of the experimental results carried out in this paper. 
\subsection{Evaluation Metrics}

As evaluation metrics, we have used area under the Receiver Operator Curve (AUCROC) and area under the Precision Recall Curve (AUPR) . These curves use Precision, Recall (TPR) and False Positive Rate (FPR) as underlying metrics . 

\begin{equation}
Precision= {T P \over T P + F P} 
\end{equation}

\begin{equation}
TPR = { TP \over T P + F N  }
\end{equation}

\begin{equation}
FPR = {  FP \over F P + T N  }
\end{equation}


Receiver Operating Characteristic (ROC) curve represents false positive rate (fpr) along the horizontal axis and true positive rate (tpr) along the vertical axis. A perfect classifier will have Area Under ROC Curve (AUROC) of 1 which means all instances of the positive class instances have been correctly classified and none of the negative class instances have been flagged as positive. AUROC provides an ideal summary of the classifier performance. For a not so good classifier TPR and  FPR increase proportionally which brings the AUROC down. A classifier which is able to correctly classify high number of both positive and negative class instances gets a high AUROC which is our goal in case of imbalanced datasets.       
AUPR represents tpr down the horizontal axis and precision down the vertical axis. Precision and TPR are inversely related, ie. as Precision increases, TPR falls and vice-versa. A balance between these two needs to be achieved by the classifier, and to achieve this and to compare performance, AUPR curve is used.

Both of the aforementioned evaluation metrics are held as benchmarks for the assessment of classifier performance on imbalanced datasets. However, AUPR is more informative for cases of high class imbalance AUROC. This is because a large change in false positive counts can result in a small change in the FPR represented in ROC. However, the same change results in a greater change of precision since it compares the false positives to the true positives instead of the true negative instances \cite{davis2006relationship}. 

\subsection{Results}

We have compared the performance of our proposed method LIUBoost against that of RUSBoost over 18 imbalanced datasets with varying imbalance ratio. All these datasets are from KEEL Dataset Repository \cite{keel}. Table~\ref{my-label} contains a brief description of these datasets.  

\begin{table}[H]
	\centering
	\caption{Dataset Description}
	\label{my-label}
	\begin{tabular}{|p{3cm}|c|c|p{1cm}|}
		\hline
		Datasets                         & Instances & Features & IR \\ \hline
		pima                             & 768       & 8        & 1.87            \\ \hline
		glass5                           & 214       & 9        & 22.78           \\ \hline
		yeast5                           & 1484      & 8        & 38.73           \\ \hline
		yeast6                           & 1484      & 8        & 41.4            \\ \hline
		ecoli-0-3-4\_vs\_5               & 200       & 7        & 9               \\ \hline
		abalone19                        & 4174      & 8        & 129.44          \\ \hline
		pageblocks                       & 548       & 10       & 164             \\ \hline
		led7digit-0-2-4-5-6-7-8-9\_vs\_1 & 443       & 7        & 10.97           \\ \hline
		glass-0-1-4-6\_vs\_2             & 205       & 9        & 11.06           \\ \hline
		glass2                           & 214       & 9        & 11.59           \\ \hline
		glass6                           & 214       & 9        & 6.38            \\ \hline
		yeast-1\_vs\_7                   & 459       & 7        & 14.3            \\ \hline
		poker-8-9\_vs\_6                 & 1485      & 10       & 58.4            \\ \hline
		haberman                         & 306       & 3        & 2.78            \\ \hline
		winequality-red-8\_vs\_6         & 656       & 11       & 35.44           \\ \hline
		glass0                           & 214       & 9        & 2.06            \\ \hline
		glass-0-1-5\_vs\_2               & 172       & 9        & 9.12            \\ \hline
		yeast-0-2-5-7-9\_vs\_3-6-8       & 1004      & 8        & 9.14			\\ \hline
	\end{tabular}
\end{table}

\begin{table}
	\caption{Wilcoxon Signed Rank Test Based on Average AUROC}
	\label{auroc_wilcoxon}	
	\centering	
	\begin{tabular}{| c | c | p{2.9cm} |c|}
		\hline
		RUSBoost &  LIUBoost &  Hypothesis
		(alpha=0.05) &p-value\\ \hline
		\\[-1em]
		11.5                         &     159.5 &                       Rejected for LIUBoost &  0.00068 \\ \hline
	\end{tabular}
\end{table}

 The algorithms have been run 30 times using 10 fold cross validation on each dataset and the average AUROC and AUROC are presented in table \ref{auroc_table} and \ref{aupr_table} respectively. Decision tree estimator C4.5 has been used as base learner. Both RUSBoost and LIUBoost have been implemented in python. All the experiments have been designed using scikit-learn \cite{scikit-learn} library.

\begin{table}
\caption{Average AUROC Comparison}
\label{auroc_table}	
\centering	
\begin{tabular}{| p{3cm} | c | c |}
	\hline
	Dataset &  RUSBoost &  Proposed Method \\ \hline
	glass5                         &     0.977 &                       \textbf{0.987} \\ \hline
	yeast5                         &     0.984 &                       \textbf{0.988} \\ \hline
	yeast6                         &     0.916 &                       \textbf{0.921} \\ \hline
	ecoli-0-3-4\_vs\_5               &     \textbf{0.987} &                      0.981 \\ \hline
	abalone19                      &     0.784 &                       \textbf{0.801} \\ \hline
	glass-0-1-4-6\_vs\_2             &     0.701 &                       \textbf{0.780} \\ \hline
	glass2                         &     0.697 &                       \textbf{0.794} \\ \hline
	page-blocks0                   &     \textbf{0.988} &                       \textbf{0.988} \\ \hline
	glass6                         &     0.961 &                       \textbf{0.966} \\ \hline
	yeast-1\_vs\_7                         &     0.785 &             \textbf{0.794} \\ \hline
	poker-8-9\_vs\_6                 &     0.791 &                       \textbf{0.792} \\ \hline
	haberman                       &     0.599 &                       \textbf{0.647} \\ \hline
	winequality-red-8\_vs\_6         &     0.708 &                       \textbf{0.727} \\ \hline
	led7digit-0-2-4-5-6-7-8-9\_vs\_1 &     0.943 &                       \textbf{0.953} \\ \hline
	glass0                         &     0.858 &                       \textbf{0.869} \\ \hline
	glass-0-1-5\_vs\_2               &     0.646 &                       \textbf{0.725} \\ \hline
	yeast-0-2-5-7-9\_vs\_3-6-8       &     \textbf{0.941} &                       0.938 \\ \hline
	pima                           &     0.689 &                       \textbf{0.704} \\ \hline
\end{tabular}
\end{table}

From the results presented in Table~\ref{auroc_table}, we can see that with respect to AUROC, LIUBoost outperformed RUSBoost over 15 datasets. However, with respect to AUPR, LIUBoost outperformed RUSBoost over 14 datasets out of 15. Results can be found in Table~\ref{aupr_table}.

\begin{table}
\caption{Average AUPR Comparison}
\label{aupr_table}	
\centering	
	\begin{tabular}{| p{3cm} | c | c |}
		\hline
		Dataset &  RUSBoost &  Proposed Method \\ \hline
		glass5                         &     0.766 &                       \textbf{0.835} \\ \hline
		yeast5                         &     0.690 &                       \textbf{0.742} \\ \hline
		yeast6                         &     0.457 &                       \textbf{0.548} \\ \hline
		ecoli-0-3-4\_vs\_5               &     \textbf{0.930} &            0.915 \\ \hline
		abalone19                      &     \textbf{0.998} &                       \textbf{0.998} \\ \hline
		glass-0-1-4-6\_vs\_2             &     0.209 &                       \textbf{0.258} \\ \hline
		glass2                         &     0.257 &                       \textbf{0.263} \\ \hline
		page-blocks0                   &     0.905 &                       \textbf{0.907} \\ \hline
		glass6                         &     0.893 &                       \textbf{0.923} \\ \hline
		yeast-1\_vs\_7                         &     \textbf{0.403} &             0.344 \\ \hline
		poker-8-9\_vs\_6                 &     0.188 &                       \textbf{0.249} \\ \hline
		haberman                       &     0.344 &                       \textbf{0.392} \\ \hline
		winequality-red-8\_vs\_6         &     0.192 &                       \textbf{0.242} \\ \hline
		led7digit-0-2-4-5-6-7-8-9\_vs\_1 &     0.648 &                       \textbf{0.759} \\ \hline
		glass0                         &     0.708 &                       \textbf{0.753} \\ \hline
		glass-0-1-5\_vs\_2               &     0.220 &                       \textbf{0.263} \\ \hline
		yeast-0-2-5-7-9\_vs\_3-6-8       &     \textbf{0.835} &                       0.824 \\ \hline
		pima                           &     0.529 &                      \textbf{ 0.544} \\ \hline
	\end{tabular}
\end{table}

We have performed Wilcoxon Pairwise Signed Rank Test\cite{wilcoxontest} in order to ensure that the improvements achieved by LIUBoost are statistically significant. This is highly recommended for comparing the performance of two machine learning algorithms.
The test results indicate that the performance improvements both with respect to aupr and auroc are significant since the null hypothesis of equal performance has been rejected at 5\% level of significance in favor of LIUBoost. Wilcoxon test results can be found in Table~\ref{auroc_wilcoxon} and Table~\ref{aupr_wilcoxon}.

\begin{table}[H]
	\caption{Wilcoxon Signed Rank Test Based on Average AUPR}
	\label{aupr_wilcoxon}	
	\centering	
	\begin{tabular}{| c | c | p{2.9cm} |c|}
		\hline
		RUSBoost &  LIUBoost &  Hypothesis
		(alpha=0.05) & p-value \\ \hline
		\\[-1em]
		23.5                         &     146.5 &                       Rejected for LIUBoost & 0.0037\\ \hline
	\end{tabular}
\end{table}

\section{Conclusion}
\label{conclusion}
In this paper, we have proposed a novel boosting based algorithm for dealing with the problem of class imbalance. Our method LIUBoost is the first one to combine both sampling technique and cost-sensitive learning. Although good number of methods have been proposed for dealing with imbalanced datasets, none of them have proposed such an approach. We have tried to design an ensemble method that would be cost-efficient just like RUSBoost but would not suffer from the resulting information loss and the results so far are satisfying. Additionally, recent research has indicated that dividing the minority class into categories is the right way to go for imbalanced datasets\cite{borowska2017rough,napierala2016types}. In our opinion, both majority and minority instances should be divided into categories and the hard instances should be given special importance in imbalanced datasets. This becomes even more important when the underlying sampling technique discards some instances for data balancing. 

Class imbalance is prevalent in many real world classification problems. However, the proposed methods have their own deficits. Cost-sensitive methods suffer from domain specific cost assignment schemes while oversampling based methods suffer from overfitting and increased runtime. Under such scenario, LIUBoost is cost-efficient, defines a generic cost assignment scheme, does not introduce any false structure and takes into account additional problems such as bad minority hubs and class overlapping. The results are also statistically significant. In future work, we would like to experiment with other cost assignment schemes.            

\bibliographystyle{IEEEtran}
\bibliography{ICCIT2017}

\end{document}